# A novel mutation operator based on the union of fitness and design spaces information for Differential Evolution

Hossein Sharifi Noghabi[1, 2], Habib Rajabi Mashhadi[1,2,3,*] and Kambiz Shojaei[1,2]

[1]Center of Excellence on Soft Computing and Intelligent Information Processing (SCIIP), Faculty of Engineering, Ferdowsi University of Mashhad, Iran.
[2]Department of Computer Engineering, Ferdowsi University of Mashhad, Mashhad, Iran.
[3]Department of Electrical Engineering, Ferdowsi University of Mashhad, Mashhad, Iran.

**Abstract**

Differential Evolution (DE) is one of the most successful and powerful evolutionary algorithms for global optimization problem. The most important operator in this algorithm is mutation operator which parents are selected randomly to participate in it. Recently, numerous papers are tried to make this operator more intelligent by selection of parents for mutation intelligently. The intelligent selection for mutation vectors is performed by applying design space (also known as decision space) criterion or fitness space criterion, however, in both cases, half of valuable information of the problem space is disregarded. In this article, a Union Differential Evolution (UDE) is proposed which takes advantage of both design and fitness spaces criteria for intelligent selection of mutation vectors. The experimental analysis on UDE are performed on CEC2005 benchmarks and the results stated that UDE significantly improved the performance of differential evolution in comparison with other methods that only use one criterion for intelligent selection.

---

[*] Corresponding Author, Department of Computer Engineering, Faculty of Engineering, Ferdowsi University of Mashhad, University Complex, Vakil Abad BLV. Azadi Sq. Mashhad, Iran. Email: h_mashhadi@um.ac.ir; hsharifi@stu.um.ac.ir; h.sharifi1990@yahoo.com

1. INTRODUCTION

In order to solve complicated computational problems, researchers and scientists are taking advantage of nature and inspiring from it for finding an intelligent, innovative and effective solutions and evolutionary computation is one of the most important results of this inspiration (Bäck, 1996), Michalewicz, 1996). In an iterative process, an evolutionary algorithm moves a single individual or a population toward the global optimum. Algorithms such as Genetic Algorithm (GA), Particle Swarm Optimization (PSO), Simulated Annealing (SA), etc. are examples of evolutionary and meta-heuristic algorithms (Thomas, David and Zbigniew, 1997), Boussaïd, Lepagnot and Siarry, 2013).

Among these methods, Differential Evolution (DE) is one of the most effective and powerful algorithms for continuous global optimization problem (Price, 1999). DE was originally proposed in 1995 by Storn and Price (Storn and Price, 1997). DE is based on population and similar to other evolutionary methods, it has some operators for handling exploration and exploitation in its optimization process. There are numerous advantages about DE including, its simplicity in implementation, a few control parameters and the fact that it is intersection of classic methods such as Nelder-Mead and Controlled Random Search and early evolutionary algorithms such as GA and SA (Price, Storn and Lampinen, 2005), Das and Suganthan, 2011), Neri and Tirronen, 2010).

The performance of DE significantly relays on two things: first, the strategy of generating new offspring which is performed by mutation and crossover operators and second, the mechanism of controlling the parameters of the algorithm including, number of population, scaling factor and crossover rate (Brest, Zamuda, Bošković, Greiner and Žumer, 2008). Generally, in the process of evolution and optimization, the first step (generating new offspring) takes place and after that parameters of the algorithm will be tuned and this steps are repeating until the algorithm reaches its stopping condition (Das, Mandal and Mukherjee, 2014), Wei-Jie, Meie, Wei-Neng, Zhi-Hui, Yue-Jiao, Ying, Ou and Jun, 2014).

In order to enhance the performance of DE, in recent years, intelligent selection of mutation vectors received a lot of attention in DE literature (Das and Suganthan, 2011), Biswas, Kundu and Das, 2014), Wenyin and

Zhihua, 2013). Usually, parent vectors in mutation are selected randomly from the population and since this strategy only deals with exploration, applying the best member of population (Price, Storn and Lampinen, 2005)., p% best (Jingqiao and Sanderson, 2009)., current members, etc. are suggested later to provide DE with exploitation as well as exploration (Matej, repin, ek, Liu and Mernik, 2013).

Intelligent selection methods generally take advantage of information in either design or fitness space (Das and Suganthan, 2011)., for instance, Kaelo and Ali applied tournament selection in fitness space (Kaelo and Ali, 2006)., Epitropakis *et al.* (Epitropakis, Tasoulis, Pavlidis, Plagianakos and Vrahatis, 2011). defined the selection criterion in design space and Gong and Cai (Wenyin and Zhihua, 2013). proposed a ranking approach in fitness space for intelligent selection for mutation vectors. Although all of these methods have their own merits, the main drawback of them is that they disregard half of the valuable information for determination of mutation vectors. In nature, good species have good information and consequently they can guide the whole population to the global optimum (Wenyin and Zhihua, 2013). Based on this fact, in this article, Union Differential Evolution (UDE) is proposed that makes the best use of information in both design and fitness spaces in mutation operator for selection of parent vectors. Among five parent vectors in the proposed mutation, two of them are selected with fitness space criterion, two of them are selected randomly and the last one is chosen based on design space criterion.

Some other DE variants use more than one mutation in similar idea of concepts like boosting or ensemble learning. Although such methods are beyond the score of this paper, a brief overview can be helpful. Self-adaptive DE, SaDE (Qin, Huang and Suganthan, 2009). uses both multiple trial vector strategies i.e. multiple mutations and at the same time a learning mechanism to update parameters associated with them. Wang *et al*. proposed Composite DE, CoDE (Yong, Zixing and Qingfu, 2011). which improved performance of DE by applying three mutation operators and three control parameter settings. Mallipeddi *et. al.* (Mallipeddi and Suganthan, 2010). proposed EPSDE which has an ensemble mutation and crossover strategies along with a pool for controlling parameters. Later, Iacca *et. al.* (Iacca, Neri, Caraffini and Suganthan, 2014). introduced EPSDE-LS with a pool of local search algorithms and proposed a new framework for ensemble DE. Sharifi

Noghabi *et. al.* (Noghabi, Mashhadi and Shojaei, 2015). proposed DE with Generalized Mutation operator, GMDE which has two mutation pools with four trial vector generating strategies in each and applied it for parameter optimization in gene selection problem (Mohammadi, Sharifi Noghabi, Abed Hodtani and Rajabi Mashhadi, 2016). More interested readers can refer to (Das, Mullick and Suganthan, 2016).

The rest of the paper is organized as follows, section II is the brief review of previous research about DE and intelligent selection, section III deals with the proposed method and section IV is about experimental results and analysis.

## 2. RELATED WORK

In this section, first, the original DE is briefly introduced and then related works regarding intelligent selection are discussed.

Without loss of generality, throughout this paper, the following numerical optimization problem is considered:

$$Minimize \quad f(x), \quad x \in CS \tag{1}$$

Where, $CS \subseteq R^D$ is a compact set, $x = [x_1, x_2, ..., x_D]^T$ and D is the dimension of the problem space. Moreover:

$$LB \leq x_j \leq UB, \quad j = 1, 2, ..., D \tag{2}$$

Where, LB and UB are lower bound and upper bound respectively.

### A. Original DE

Similar to other evolutionary computation methods, DE starts with an initial population that generally initialized randomly. After determining the population, a new candidate individual is generated by applying mutation and crossover operators (Price, Storn and Lampinen, 2005). This candidate then becomes the input of selection operator and through a hard selection mechanism between the candidate and the current member of the population, if the candidate is better than the current member it will enter the next generation otherwise the current member remains in the population (Das and Suganthan, 2011). Algorithm 1 is the pseudocode of the original DE. In this algorithm, *NP* is the population size, *D* is size of dimension, *F* is the scaling factor, *CR* is the crossover rate, *randint (1, D)* is a random integer in [1, *D*] and *rand* is a random real number in [0, 1].

Algorithm 1 original DE for DE/rand/1/bin

1: Generate an initial population consisting of *NP* individuals and evaluate them.

2: **While** (termination criterion is not satisfied)

3:     **For** $i = 1: NP$

4:         Select randomly $r_1 \neq r_2 \neq r_3 \neq i$

5:         $j_{rand} = randint (1, D)$

6:             **For** $j = 1:D$

7:                 **If** $rand < CR \;||\; j = j_{rand}$ **then**

8:                     $u_{i,j} = x_{r1,j} + F \cdot (x_{r2,j} - x_{r3,j})$

9:                 **Else**

10:                    $u_{i,j} = x_{i,j}$

11:                **End if**

12:            **End for**

13:    **End for**

14:    **For** $i = 1 : Np$ **do**

16:        Evaluate the offspring $u_i$

17:        **if** $f(u_i) < f(x_i)$ **then**

18:            Replace $x_i$ with $u_i$

19:        **End if**

20:    **End for**

21: **End while**

*B. Mutation Operators in DE*

The main operator in DE is mutation operator and a lot of mutations have been proposed in DE literature from 1997 up to now (Das and Suganthan, 2011), Neri and Tirronen, 2010). These mutations take advantage of diverse mechanisms and strategies for generating the *donor* vector. In order to distinguish between DE variants, notation DE/X/Y/Z proposed by Storn and Price where DE denotes Differential Evolution, X denotes the base vector, Y denotes number of difference vectors and finally Z determines the type of crossover (Storn and Price, 1997). In this paper, since in all of the cases, the binomial crossover (bin) is applied, therefore, Z is omitted from the notation.

Some of the well-known mutations are listed as follows:

DE/rand/1

$$X_{new,g} = X_{r1,g} + F.(X_{r2,g} - X_{r3,g}) \tag{3}$$

DE/best/1

$$X_{new,g} = X_{best} + F.(X_{r1,g} - X_{r2,g}) \tag{4}$$

DE/rand/2

$$X_{new,g} = X_{r1,g} + F.(X_{r2,g} - X_{r3,g}) + F.(X_{r4,g} - X_{r5,g}) \tag{5}$$

DE/best/2

$$X_{new,g} = X_{best} + F.(X_{r1,g} - X_{r2,g}) + F.(X_{r3,g} - X_{r4,g}) \tag{6}$$

DE/current-to-best/1

$$X_{new,g} = X_{current,g} + F.(X_{best} - X_{current,g}) + F.(X_{r2,g} - X_{r3,g}) \tag{7}$$

DE/rand-to-best/1

$$X_{new,g} = X_{r1,g} + F.(X_{best} - X_{r1,g}) + F.(X_{r2,g} - X_{r3,g}) \tag{8}$$

DE/current-to-rand/1  $(F' = ki \cdot F)$

$$X_{new,g} = X_{current,g} + k.(X_{r1,g} - X_{current,g}) + F'.(X_{r2,g} - X_{r3,g}) \qquad (9)$$

Where, in these mutations, $X_{new,g}$ is the newly generated member, $X_{best}$ indicates the best individual, $X_{current}$ denotes the current member of the population at generation $g$, $Xr_1 \neq Xr_2 \neq Xr_3 \neq Xr_4 \neq Xr_5 \neq X_{current}$ and $F$ is the scaling factor. $K$ is a constant weight between the engaged terms.

*C. Intelligent selection for mutation*

Intelligent selection has been under consideration since the original DE was proposed, in fact, some methods such as DE/best1 and DE/current-to-best/1 can be considered examples of this category of mutations. Although Das *et al*. (Das and Suganthan, 2011). introduced intelligent selection as one of the promising area in DE, more serious and deep studies started from 2006 by Kaelo and Ali (Kaelo and Ali, 2006)., they proposed DERL which applies tournament selection to select three random member of population and then the best of them is placed as the base vector and the other two as difference vectors in DE/rand/1 mutation. Jingqia *et al*. (Jingqiao and Sanderson, 2009). proposed DE/current-to-*p*best/1 in JADE that uses p% of the best members of population and moreover; it also applies an archive of failed answers to participate in mutation for more diversity. The same approach also introduced in (Baatar, Dianhai and Chang-Seop, 2013).. Epitropakis *et al.* (Epitropakis, Tasoulis, Pavlidis, Plagianakos and Vrahatis, 2011). presented a proximity-based mutation that takes advantage of design space criterion for intelligent selection. In this method, first Affinity matrix (*NP\*NP*) is calculated based on Euclidean distance of individuals from each other and then the selection probabilities are calculated. The closer an individual to the current member is, the more probability it has. Finally, on each row of the probabilities matrix individuals are chosen by roulette wheel selection. In (García-Martínez, Rodríguez and Lozano, 2011)., Garcia *et al.* proposed the role differentiation and malleable mating for mutation. In this approach, the vectors in the population are grouped into four groups, including *receiving*, *placing*, *leading*, and *correcting* groups. For mutation and crossover operators, vectors are selected from corresponding groups, instead of the whole population. Gong and Cai in (Wenyin and Zhihua, 2013). presented Ranking-based mutation which uses fitness space as the criterion for intelligent selection. In Ranking, first individuals are sorted and then each individual receives a rank based on its fitness. The probability of selection for each individual is calculated according to its corresponding rank in the population. For a mutation such as DE/rand/1 two of the three vectors are chosen by ranking criterion via a proportional selection method. In (Liang, Qu, Mao, Niu and Wang, 2014).. Liang *et al.* proposed an intelligent selection with fitness Euclidean-distance ratio (FER) for multimodal optimization that probabilities of selection are calculated based on this measure.

### 3. PROPOSED METHOD

In this paper, we proposed a novel mutation operator, Union DE that has both merits of design and fitness spaces in parent vectors. Since good individuals in the population have valuable information, they can guide the rest of the population much better, thus, it is important to use valuable information of both spaces. In this section, first the proposed mutation, UDE is introduced and then some definitions and criteria are stated regarding it.

#### A. UDE

In the spirit of DE/rand/2 mutation, UDE is defined as follows:

$$v_i = X_{FS1} + F.(X_{FS2} - X_{r1}) + F.(X_{DS} - X_{r2}) \quad (10)$$

Where, $v_i$ is the donor vector, $F$ is the scaling factor, $X_{FSi}$ is the parent vector that chosen by fitness space criterion, $X_{DS}$ is the vector that selected by design space criterion and $X_{ri}$ is a randomly selected vector.

Terminology 1: *Base role:* a vector has base role if it is in the first position of mutation operator. For example, $X_{FS1}$ has the base role in Eq. (10).

Terminology 2: *Leading role*: a vector has leading role if it is in the first place of a difference vector. In Eq. (10), $X_{FS2}$ and $X_{DS}$ have this role.

Terminology 3: *Terminal role*: a vector has terminal role if it is in the second position of a difference vector. For example, $X_{r1}$ and $X_{r2}$ in Eq. (10) are terminal vectors.

***Remark 1.*** In contrast with Ranking and Proximity mutations, UDE takes advantage of intelligent selection for determination of base and leading roles in mutation and this approach provides the algorithm with better performance.

*B. Fitness space criterion*

In order to select $X_{FS1}$ and $X_{FS2}$ vectors, following steps are required:

First sort the population in increasing order (from best to worst) according to their fitness value.

Second, calculate the selection probability for each individual by:

$$P_i = \frac{NP - i}{NP}, i = 1, 2, ..., NP \tag{11}$$

Where, $P_i$ is the selection probability for the *i-th* member and *NP* is number of population.

After measuring selection probabilities, two members are chosen by roulette wheel and randomly assign to stated roles for fitness space vectors. Therefore, the criterion for fitness space is calculated.

*C. Design space criterion*

For design space selection, based on Euclidean distance between all of the individuals in the population, the *Distance Matrix*, *DM* is obtained as follows:

$$DM = \begin{pmatrix} \|X_1 - X_1\| & \cdots & \|X_1 - X_{NP}\| \\ \vdots & \ddots & \vdots \\ \|X_{NP} - X_1\| & \cdots & \|X_{NP} - X_{NP}\| \end{pmatrix} \tag{12}$$

Due to symmetric property of distance, *DM* is a symmetric and only its upper triangular part requires to be calculated. Based on *DM*, we calculate the *Probability Matrix*, *PM* by the following equation:

$$PM(i,j) = 1 - \frac{DM(i,j)}{\sum_k DM(i,k)} \tag{13}$$

In this matrix, *PM (i, j)* represents the probability between *i-th* and *j-th* individual with respect to the *i-th* row of PM. This probability is inversely proportional to the distance of the *j-th* member of the population which means the member with minimum distance to *i-th* member will have the maximum probability. In order to choose $X_{DS}$, roulette wheel selection without replacement is performed on every row of *PM* matrix (for each member of population).

***Remark 2.*** Whenever a new member is entered to the population only row and column associated to this new individual are required to be updated and it is not necessary to calculate the entire matrices.

***Remark 3.*** Applying Eq. (13) can be considered as a local search as well because it exploits the regions around a pre-defined member by assigning the higher probability to closer individuals. Hence, this approach makes UDE competitive for multimodal problems too. Algorithm 2 is pseudocode for UDE.

Algorithm 2 Union DE (UDE)

1: Generate an initial population consisting of *NP* individuals and evaluate them.

2: **While** (termination criterion is not satisfied)

3:     **For** *i = 1: NP*

4:         Select randomly $r_1 \neq r_2 \neq i$

5:     select $X_{FS1}$ and $X_{FS2}$ according to Eq. (11) and roulette wheel.

7:     select $X_{DS}$ according to Eqs. (12-13) and roulette wheel.

8:         Perform UDE mutation according to Eq. (10).

9:         Perform binomial crossover similar to Algorithm 1.

10:    **End for**

11:    **For** *i = 1 : Np* **do**

12:        Evaluate the offspring $u_i$

13:        **if** *f*(*ui*) < *f*(*xi*) **then**

14:            Replace *xi* with *ui*

15:        **End if**

16:    **End for**

17: **End while**

## 4. EXPERIMENTAL RESULTS AND ANALYSIS

In this section, in order to investigate and study the performance of the proposed UDE, comprehensive experiments are performed. We applied 25 benchmark functions from CEC2005 competition (P. N. Suganthan, N. Hansen, J. J. Liang, K. Deb, A. Auger and Tiwari, 2005).. These functions are categorized into four groups: 1) F1-F5: these functions are unimodal, 2) F5-F12: these functions are basic multimodal functions, 3) F13 and F14 are expanded multimodal functions and finally 4) F15-F25 are hybrid composition functions.

### A. Parameter settings

Since the value of parameters is totally influential on the performance of the algorithm, in all cases, we utilized jDE control parameter mechanism to have fair situation for all of the algorithms. jDE (Brest, Greiner, Boskovic, Mernik and Zumer, 2006). controls scaling factor and crossover rate as follows:

$$F_{i,G+1} = \begin{cases} F_l + rand_1 * F_u & rand_2 < \tau_1 \\ F_{i,G} & othewise \end{cases} \quad (14)$$

$$Cr_{i,G+1} = \begin{cases} rand_3 & rand_4 < \tau_2 \\ Cr_{i,G} & otherwise \end{cases} \quad (15)$$

Where, $rand_{i,\, i \in 1...4} = \{rand \in [0,1]\}, \tau_1 = \tau_2 = 0.1, F_l = 0.1, F_u = 0.9$

This approach makes $F \in [0.1, 0.9]$ and $Cr \in [0,1]$.

The other parameters are initialized as follows:

NP=50; Maximum run=50; dimensions=30 and maximum number of function evaluation=D * 10 000.

*B. Statistical analysis*

In this paper, we used Wilcoxon signed-rank test at $\alpha = 0.05$ for comparing proposed UDE with other methods. According to this analysis, *win* denotes number of functions that UDE performed better than the compared method, *lose* denotes number of functions that UDE was not successful compared to the corresponding method and *tie* denotes number of functions that UDE and the compared method are equal.

*C. Results and experiments*

In order to study and performance of the presented UDE, we selected DERL, Proximity-based and Ranking based mutations to compare with. All of these methods take advantage of DE/rand/2 mutation that developed by their own approaches. Each method executed 50 times independently for the stated settings and the average value of the errors for each function is presented in TABLE 1. In this table, boldface indicates that for the corresponding function that algorithm achieved the best result among the other methods. According to this table, with respect to other methods, UDE significantly overcomes the other methods in most of the cases. Compared to DERL, UDE wins in 16 functions, ties in 2 and lost in 7 functions which most of successes of DERL were in unimodal and multimodal functions. UDE also wins in 15 functions in comparison with Proximity-based mutation, ties in 3 and lost in 7 functions similar to DERL. In comparison with Ranking-based mutation, UDE again wins in 16 cases and ties in 4 and only loses in 5 functions. In case of proximity and ranking, they appeared to be more competitive in hybrid functions (F15-F25). From another point of view, in functions 1 to 5 (unimodals) in three of them including F1, F4 and F5 UDE had the best answer among the others, ranking find the best answer in one of the cases (F1) and DERL was the best in two among the other methods (F2 and F3), thus, for unimodal functions, utilizing both design and fitness spaces leads to better performance. For basic multimodal functions, ranking had the best performance in two functions including F7 and F9, proximity found the best answer in F9 too and DERL only found the best answer in F12. For this functions again UDE had the best performance and obtained the best answer in five functions (F6, F8, F9, F10, and F11). Hence, for this group, UDE with its hybrid approach for intelligent selection tremendously overcomes the other methods. For DERL and proximity were better than other methods and won in F14 and F13 respectively. In case of hybrid composition functions, all of the methods performed the same for F24, so, beside this function, DERL was successful only in F25 ranking achieved the best answer among others in F15, F21 and F23. However, proximity performed generally better in comparison with ranking and DERL and found the best answer in F17, F18 and F19. Although both ranking and proximity relatively performed well for hybrid composition functions especially proximity, UDE again obtained the best results among other methods and found the best answer in F16, F20, F21 and F22. In addition to its superiority, for rest of the functions and in most of them, UDE achieved to answers that were extremely close to the best ever found for that function. Therefore, hybrid approach for intelligent selection significantly improved the performance of the algorithm for basic multimodal and hybrid composition functions. Although UDE found relatively adequate minimums, it did not act well enough in expanded multimodal functions. It also appeared for complex functions such as F15-F25, after hybrid approach presented in UDE, design space criterion (proximity) performed better than fitness space approach.

In summary, among 24 functions (without considering F24) DERL found the best answer in five of the cases, proximity performs the same as DERL and found the best answer in five functions but statistically, proximity performs significantly better than DERL. Ranking found the best in six functions but, statistically, there is no meaningful difference between proximity and ranking. UDE achieved the best performance in 12 functions and more importantly, statistical analysis regarding its performance indicates that in all of comparisons, UDE is better than the other methods and there is significant and meaningful difference between them in favor of UDE. TABLE 2 is presented the statistical details regarding UDE. In this table, MR- indicates mean of negative ranks, MR+ indicates mean of positive ranks, SR- denotes sum of negative ranks, SR+ denotes sum of positive ranks. P-value is the measure that determines the difference between algorithms is significant or not and finally in the last column, "+" indicates that UDE is significantly better than the compared method, "-" indicates that UDE is significantly worse than the compared method and "=" indicates that there is no meaningful difference between UDE and the corresponding method.

As illustrated in this table, in all of the cases, UDE is significantly better than the other algorithms which means that there is a meaningful difference between the proposed method and the other algorithms in favor of UDE. Therefore, the proposed method, make the best use of good information in design and fitness spaces and this usage is leading to an effective, robust and accurate global optimization method which is capable of working with diverse kinds of problems.

*D. Comparison with two basic mutations*

UDE might be considered similar to DE/best/2 in a way. Therefore, one interesting study can be comparison of the proposed method with such a basic mutation. Results for this study are tabulated in Table 3 and as expected, UDE had significantly better performance than DE/best/2 method in most of the cases. In order to save space and avoid unnecessary large tables, we only report the results of the statistical analysis for this study.

## 5. CONCLUSION AND FUTURE WORKS

In this paper, we studied the impact of intelligent selection for mutation operator. Intelligent selection is a new generation in DE literature which is trying to instead of choosing parent vectors randomly, select them intelligently. Current methods utilize design space or fitness space criteria, however, the consequent of such a selection is losing half of the valuable information available in the problem space. In this paper, we proposed Union Differential Evolution (UDE) which takes advantage of both stated spaces in a single mutation strategy. In the presented mutation, there are five vectors, two of them are selected according to fitness space criterion, another two vectors are selected randomly and the last one is chosen by design space criterion, therefore, both exploration and exploitation are provided in UDE.

The results are illustrated that UDE is significantly better than methods that solely use fitness or design space criteria in their evolution process. All extensive experiments were performed on 25 benchmark functions from CEC2005 competition.

In future, we intend to study the impact of population sizes on the performance of the proposed method, more importantly, the proposed method should be examined with other control parameter approaches. Moreover, it is essential to investigate the performance of the proposed method for higher dimensions as well.

## ACKNOWLEDGMENT

The authors would like to thank Center of Excellence on Soft Computing and Intelligent Information Processing (SCIIP) for kind supports and Dr. Y. Wang for making the MATLAB code of jDE available online.

## CONFLICT OF INTERESTS

Authors have nothing to disclose regarding conflict of interests.

TABLE 1 COMPARISON OF THE PERFORMANCE
BETWEEN UDE, DERL, PEOXIMITY AND RANKING
FOR FUNCTIONS F1–F25 AT D = 30 AND 300 000
FUNCTION EVALUATION FOR 50 INDEPENDENT RUNS

| Fun. | DERL mean | Proximity mean | Ranking mean | UDE mean |
|---|---|---|---|---|
| F1 | 3.26E-28 | 5.05E-31 | **0** | **0** |
| F2 | **1.31E-23** | 3.61E-10 | 0.0162176 | 2.92E-10 |
| F3 | **74543.26** | 164945.8 | 379023.67 | 142905.61 |
| F4 | 254.11604 | 0.8932487 | 8.6211149 | **0.013536** |
| F5 | 1806.9985 | 476.52098 | 1043.8789 | **278.8043** |
| F6 | 13.329272 | 0.5722474 | 34.920037 | **0.318399** |
| F7 | 4696.2886 | 4696.2886 | **4692.478** | 4696.2886 |
| F8 | 20.93327 | 20.935417 | 20.954167 | **20.91975** |
| F9 | 4.7360039 | **0** | **0** | **0** |
| F10 | 56.23388 | 49.691529 | 58.571746 | **46.94091** |
| F11 | 23.833349 | 27.718178 | 27.698503 | **27.41882** |
| F12 | **12518.51** | 16496.466 | 16000.211 | 14155.216 |
| F13 | 1.399354 | **1.359917** | 1.3732692 | 1.3777869 |
| F14 | **12.23211** | 12.912602 | 12.948416 | 12.910522 |
| F15 | 370.1511 | 352.17827 | **242.8361** | 376 |
| F16 | 137.22539 | 69.01356 | 103.23722 | **68.16309** |
| F17 | 154.06222 | **133.9956** | 152.5687 | 148.4484 |
| F18 | 911.45675 | **904.5121** | 904.78486 | 904.68634 |
| F19 | 912.80443 | **904.5984** | 904.67019 | 904.71551 |
| F20 | 912.33533 | 904.94622 | 904.81682 | **904.4761** |
| F21 | 539.8539 | 512.00004 | **500** | **500** |
| F22 | 891.98448 | 881.90812 | 878.71601 | **874.3584** |
| F23 | 589.56758 | 534.16419 | **534.1641** | 534.16425 |
| F24 | **200** | **200** | **200** | **200** |
| F25 | **1617.594** | 1626.6076 | 1631.2113 | 1626.8149 |
| | Win: 16 lose: 7 tie: 2 | Win: 15 lose: 7 tie: 3 | Win: 16 lose: 5 tie: 4 | |

TABLE 2 RESULTS OF WILCOXON TEST AT 95% FOR F1-F25 AT D=30 FOR 300 000 FUNCTION EVALUATION

| Algorithm | MR- | MR+ | SR- | SR+ | P-value | Difference |
|---|---|---|---|---|---|---|
| DERL Vs. UDE | 12.25 | 11.43 | 196.00 | 80.00 | 0.039 | + |
| Proximity Vs. UDE | 12.20 | 10.00 | 183.00 | 70.00 | 0.032 | + |
| Ranking Vs. UDE | 12.13 | 7.40 | 194.00 | 37.00 | 0.006 | + |

TABLE 3 RESULTS OF WILCOXON TEST AT 95% FOR F1-F25 AT D=30 FOR 300 000 FUNCTION EVALUATION AGAINST DE/BEST/2

| Algorithm | MR- | MR+ | SR- | SR+ | P-value | Difference |
|---|---|---|---|---|---|---|
| DE/Best/2 Vs. UDE | 14.30 | 7.80 | 286.00 | 39.00 | 0.001 | + |
| Win | 20 | | | | | |
| Lose | 5 | | | | | |
| Tie | 0 | | | | | |